\documentclass[upint,subscriptcorrection,varvw,mathalfa=cal=euler,pdf-a,balance,fontspec,nofoot]{asmeconf} % <=== remove colorlinks before your *final* submission to ASME! But keep it for other purposes.
\usepackage{stfloats}
\usepackage{multirow}
\usepackage{float}
\usepackage{algorithm}
\usepackage{verbatim}
\usepackage{xcolor}

\hypersetup{%
	pdftitle={Graph Learning based Generative Design for Interdependent Network Systems},                  % <=== change to YOUR pdf file title
	pdfkeywords={Generative design, power systems, network systems, graph nueral network},% <=== change to YOUR pdf keywords
	pdfauthor={Jiaxin Wu},                          		      % <=== change to YOUR name[s]!
}

\begin{document}
% Change these fields to the right content for your conference.
% You can comment these out if for some reason you don't want a header.
% Use title case (first letters capitalized), not all capitals

% \ConfName{Proceedings of the ASME 2022\linebreak International Design Engineering Technical Conferences \&\linebreak Computers and Information in Engineering Conference}
% \ConfAcronym{IDETC/CIE 2022}
% \ConfDate{August 14-17, 2021}
% \ConfCity{St. Louis, MO, USA}
% \PaperNo{DETC2022-89622}

% Units of measure and other specialty lowercase terms in the title should be enclosed in \NoCaseChange{...} to maintain lower case type LaTeX will automatically set the rest of the title in all capital letters.

\title{Graph Learning based Generative Design for Resilience of Interdependent Network Systems} % <=== replace with YOUR title
%\title{Place Title Here: Place Subtitle After Colon} 
 
%   Put author names into the order you want. Use the same order for affiliations.
%   \affil{#} tags the author's affiliation to the address in \SetAffiliation{#}.
%   No space between last name and \affil{#}, separate names with commas.
%
%   \CorrespondingAuthor{email} follows that author's affiliation, no spaces.  
%   If multiple corresponding authors, put both email addresses in the same command and place after both authors.
%
%   \JointFirstAuthor, if applicable, follows the affiliation of the relevant authors, no spaces.

\SetAuthors{Jiaxin\ Wu\affil{1}and Pingfeng\ Wang\affil{1}\CorrespondingAuthor{pingfeng@illinois.edu}}
\SetAffiliation{1}{Department of Industrial and Enterprise Systems Engineering \\
University of Illinois Urbana-Champaign, Urbana, IL, USA\\}
%\SetAffiliation{2}{Mechanical Engineering, University of Texas Dallas, Richardson, TX, USA\\}
%	To instead use the [authorgrid] option, follow the syntax in Appendix B.

\maketitle

%%% Use this footnote for tracking various versions of your draft. Change text to suit your own needs. 
%%% REMOVE from final version of your paper.
%%% \date{..} calls the same command. 

\versionfootnote{Draft submit to Journal of Mechanical Design: version \versionno, \today}% <=== Delete before final submission.

%%% Change these to your keywords.  Keywords are automatically printed at the end of the abstract.
%%% This command MUST COME BEFORE the end of the abstract.
%%% If you don't want keywords, leave the argument of \keywords{} empty (or use the abstract* environment)


%%%%%%%%%  Abstract  %%%%%%%%%%%%%%%%%%%%%%%%%%%%%%%%%
%%
%% Abstract should be no more than 200 words
\begin{abstract}
Interconnected complex systems usually undergo disruptions due to internal uncertainties and external negative impacts such as those caused by harsh operating environments or regional natural disaster events. To maintain the operation of interconnected network systems under both internal and external challenges, design for resilience research has been conducted from both enhancing the reliability of the system through better designs and improving the failure recovery capabilities. As for enhancing the designs, challenges have arisen for designing a robust system due to the increasing scale of modern systems and the complicated underlying physical constraints. To tackle these challenges and design a resilient system efficiently, this study presents a generative design method that utilizes graph learning algorithms. The generative design framework contains a performance estimator and a candidate design generator. The generator can intelligently mine good properties from existing systems and output new designs that meet predefined performance criteria. While the estimator can efficiently predict the performance of the generated design for a fast iterative learning process. Case studies results based on power systems from the IEEE dataset have illustrated the applicability of the proposed method for designing resilient interconnected systems.
\end{abstract}

%%%%%%%%%  NOMENCLATURE %%%%%%%%%%%%%%%%%%%%%%%%%%%%%%%%%

%% To change space between the symbols and  definitions, use \begin{nomenclature}[Xcm] where X is a number 
%% The unit cm can be replaced by any LaTeX unit of dimension: pt, in, ex, em, pc, etc.
%% Default is 2em.

%% \EntryHeading{Greek letters} produces an italicized subheading, e.g., \EntryHeading{Greek letters}
%% (Leaving off the second argument of \entry, e.g. \entry{Greek letters}, has the same effect)

\begin{nomenclature}[3cm]
\EntryHeading{Parameters}
\entry{$\mathcal{G}$}{Graph}
\entry{$i,j$}{Node index in the graph}
\entry{$\mathbf{X}_v$}{Feature vector for node $v$}
\entry{$\mathbf{z}_v$}{Latent feature vector for node $v$}
\entry{$\tau_v$}{Class of node $v$}
\entry{$h_i$}{node representation in the generator}
\entry{$H$}{graph representation in the generator}
\entry{$\phi_{ij}$}{feature vector for edge generation}
%\subsection*{Variables}
\EntryHeading{Variables}
\entry{$C, L_\ell, f_1, f_2$}{Trainable neural networks}%\end{nomenclature}
\end{nomenclature}

%%%%%%%%%  BODY OF PAPER %%%%%%%%%%%%%%%%%%%%%%%%%%%%%%%%%

%%%%%%%%%%%%%%%%%%%%%%%%%%%%%%%%%%%%%%%%%%%%%%%%%%%%%%%%%%%%%%%%%%
%% 1. rewrite the estimator section: only keep the first few introduction paragraphs, and skip the detail derivation of the GCN, only need to list out the essential propagation rule
%% 2. discuss the post-disruption process, for more complex resilience evaluation, like the system actual performance toward uncertain events
%% 3. discuss about the applicability for other type of systems, e.g. simple supply chain network (~10nodes) for reproducibility (for journals)
%% 5. keep the large scale power system case for practical consideration
%% 6. There is always an issue of failures because of unforeseen, unmodelable events. How well does the method address these? What about Byzantine failures?
%% check the language to avoid using WE too much
%%%%%%%%%%%%%%%%%%%%%%%%%%%%%%%%%%%%%%%%%%%%%%%%%%%%%%%%%%%%%%%%%%

\section{INTRODUCTION}\label{sec:intro}
\begin{figure*}
    \centering
    \includegraphics[width = 0.75\linewidth]{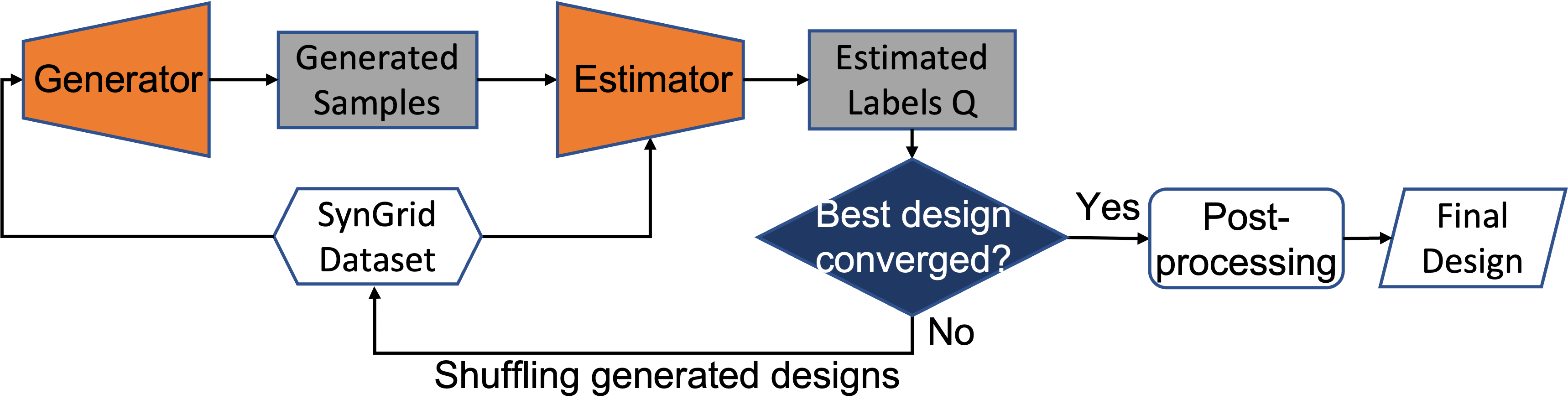}
    \caption{Flow chart of the proposed generative design framework for ICIs: there are two major components, the design generator and estimator. And both components are formulated based on graph learning algorithms.}
    \label{fig:main_flowchart}
\end{figure*}
With the increases in both scale and complexity, interdependent critical infrastructures (ICIs), such as power systems or transportation networks, become more vulnerable to disruptive events. And natural disasters impose great negative impacts on the system reliability, e.g. a winter storm and associated cold waves led to a large-scale blackout event that affected more than five million people in February 2021 in Texas \cite{Sullivan2021}. Such vulnerability, therefore, drives the research efforts that could lead to robust and resilient ICIs. For instance, how to efficiently design a large-scale system that can resist potential external disruptions, or how can the decision-maker evaluate the uncertain dynamic behavior of the ICI undergoing different disruptive events? To quantify the system's performance during disruption, or to comprehend the system's capability toward uncertain disruptive scenarios, researchers have adopted the term "resilience" from the ecology field \cite{DeAngelis1980}. Different from the terminology of system reliability, in which the time-dependent degraded system performance and the possibility of failure are studied, the resilience metric is utilized to complement the analysis of real-time system behavior. Based on the U.S. Department of Defense report, a resilient ICI should not only withstand the impacts of disruptive events but also need to acquire the capability of self-healing from damages \cite{Goerger2014}. Thus, to realize a resilient ICI through design or operational management strategies, the stakeholders need to tackle challenges in three folds: (1) how should the system proactively detect the occurrence of abnormalities with the possible external or internal disruptions; (2) how large is the bandwidth of the ICI for withstanding adversarial impacts; (3) how quick the ICI can self-recover to its nominal state \cite{Walker2004}.

Motivated by the challenges from those three aspects, different frameworks have been proposed to help the ICI establish self-healing capability after system disruptions, therefore, achieving failure resilience. Here, we categorize the research efforts about engineering resilience based on the temporal stages of the proposed frameworks, i.e. before and after the disruptions. During the post-disruption stage, several real-time operational frameworks have been proposed to guide how the system should behave after disruptive events. For instance, researchers try to attain a resilient operational framework by scheduling optimal repair tasks under uncertainties \cite{Ouyang2017, Wu2020, Wu2021a} as well as repair resources \cite{wu2021}, forming self-sustainable microgrids \cite{Chen2016, Ambia2021}, and guided recovery through control strategies \cite{DallAnese2014, Wu2019a, Wu2019}. All aforementioned studies focus on solving the optimal decisions of how to utilize the existing resources or back-ups to recover the ICI, on the promise that set up a contingency plan such as network reconfiguration beforehand. In other words, during the post disruption stage, the self-recovery capability is realized in two steps: appropriate emergency response e.g. system reconfiguration, followed by performing optimal restorations.

Although comprehensive post-disruption frameworks have been proposed to guide how a system should behave after failure events, methodologies for proactively improving the system resilience or quantifying the resilience level of the ICIs are still unknown. And without an appropriate pre-disruption design/planning framework, the stakeholders need to frequently apply the aforementioned contingency plans to ensure system nominal performance, which leads to a more significant cost for the resilience enhancement. As a result, it's required to study suitable strategies to ensure system resilience even during the planning stage and to ease the necessity of adopting post-disruption control efforts.

To improve the system resilience and thus achieve system operations with better quality, however, current engineering resilience design research has been focusing on proposing ad-hoc models. For instance, researchers have proposed different system modeling and analysis methods to quantify and analyze the resilience level of complex engineering systems, e.g. power distribution systems and supply chain networks, undergoing a disruptive event during the pre-disruption stage \cite{Yodo2016b,Sharma2017a}. Besides, various approaches based on mathematical programming models are proposed to solve the best design strategies for ICIs under different operating scenarios \cite{Bourennani2015, Li2021}. Furthermore, probabilistic approaches such as the Bayesian network has been adopted to analyze and quantify the overall system resilience with the presence of disruptions \citep{Yodo2016, Yodo2017}. All aforementioned research have demonstrated their capabilities of quantifying the system resilience and thus finding the best design of the system, e.g. expanding the existing ICI or solving the best layout for a new system. Yet existing methods have their drawbacks: either the model is ad-hoc so that the generalization is not straightforward, or the system models are simplified to have a tractable problem. For example, many optimization-based approaches have assumptions for the underlining physical constraints to derive a solvable model for optimum. Or probabilistic approaches only consider systems with around 30 components to simplify the solving process. As a result, those drawbacks limit the applicability of the existing methods on realistic large-scale ICIs.

To further complicate the pre-disruption design problems for ICIs, the system usually consists of heterogeneous components for satisfying the various needs in the same system. However, these different components add more constraints for solving the optimal design and make the decision-maker use ad-hoc models to evaluate different online scenarios. Moreover, ICIs are usually discrete systems with up to tens of thousands of nodes and edges. Such a highly combinatorial system is nearly impossible to find the best design, without the help of intelligent methods. Thus, in this study, we focus on proposing a robust and intelligent generative framework to realize the design for ICIs that optimizes the resilience level.

The proposed generative design framework for ICIs is a model-free, data-driven method. It utilizes the graph learning algorithm to reconstruct candidate designs based on the input dataset of real-world ICIs. As a result, there is no need to make any assumptions about the ICIs to simplify the system operation constraints. Also, the optimal design does not rely on any specific mathematical model of the ICI. That is, the generative approach can learn the insights from the existing ICIs directly. And throughout the iterative training process, the generated candidates can be optimized towards pre-defined performance criteria, i.e. the system resilience level. As shown in \autoref{fig:main_flowchart}, the framework includes two major components--the generator and the estimator. The design generator is a variational autoencoder (VAE) that directly outputs feasible designs for the ICI, while the estimator is pre-trained to predict the performance of the candidate design efficiently. And an iterative process that blends the generated designs into the training dataset closes the gap of “target” driven generation to bias the generator towards outputting samples with high resilience. Finally adequate post-processing step including more expensive post-disruption simulations pinpoints the best system design.

The contributions of this work are in two aspects: (1) to the authors' best knowledge, there is no “target” (resilience) driven generative algorithm for interconnected systems yet. In the machine learning community, various algorithms, such as graph recurrent neural network (GRNN) \cite{You2018}, graph recurrent attention network (GRAN) \cite{Liao2019}, and graph variational autoencoder (GVAE) \cite{Liu2018} have been proposed to generate synthetic graph/discrete structures. And with the help of the graph neural network (GNN) model, an estimator can be constructed to determine the resilience level of an ICI automatically. Combined with the generative part, the estimator further guides the generation process. (2) It's the first time adopting graph generation algorithms to physical, real-world network systems. A physical system, for instance, the power grid,  has much more complicated operational constraints than a social network or molecular structure. Those synthetic systems are the main applications of the generative algorithms proposed in the machine learning field. Thus, how to properly address the physical constraints and information when applying the graph algorithm to ICIs remains as a crucial challenge.

The rest of the paper is organized as follows: \autoref{sec:generator} explains the modeling of the design generator, and \autoref{sec:estimator} presents the formulation of the design estimator in detail. \autoref{sec:postprocessing} discusses the post-processing simulation step for re-evaluating the designs with more expensive metrics. A case study based on the power system design is used to illustrate the applicability of the proposed framework in \autoref{sec:result}. \autoref{sec:conclusion} concludes the study with brief discussions on the effectiveness of the developed design methodology.

\section{Modeling of ICIs and Resilience}
\begin{figure}
    \centering
    \includegraphics[width = \linewidth]{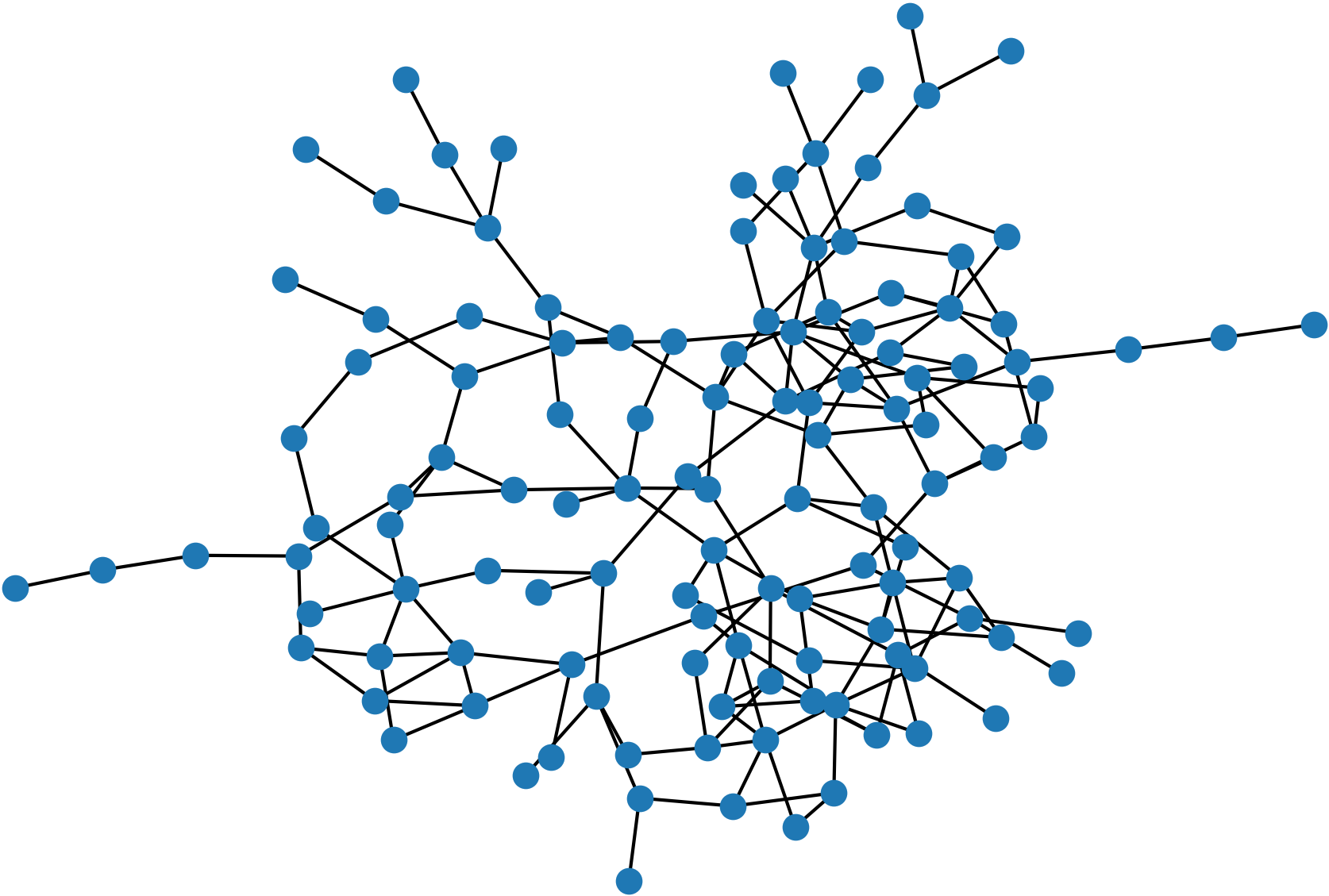}
    \caption{The graph representation of a power grid consists of 123 buses and 180 distribution lines}
    \label{fig:sample_network}
\end{figure}
In this study, the ICI is modeled as graphs to indicate the interdependence and inherent network structure of the system. Take the commodity distribution system as an example, the warehouse, transportation hubs, or the final destinations can be modeled as nodes $\{i | i\in \mathbf{V}\}$. And the distribution paths are the edges $\{ij | ij\in \mathbf{E}\}$ between different nodes. Thus, the overall ICI is denoted as a graph $\mathcal{G}:=(\mathbf{V},\mathbf{E})$ as shown in \autoref{fig:sample_network}. Other than the topological information illustrated in the figure, physical systems usually carry much more signals. Take the power grid as an example, each node of the graph can have a specific level of operating voltages, frequencies, and magnitudes of generations/loads. While the edges can have physical constraints for capacities, distance, connection types etc. Thus how to generatively design an ICI satisfying practical operating conditions becomes the major challenge.

\begin{figure}
    \centering
    \includegraphics[width = \linewidth]{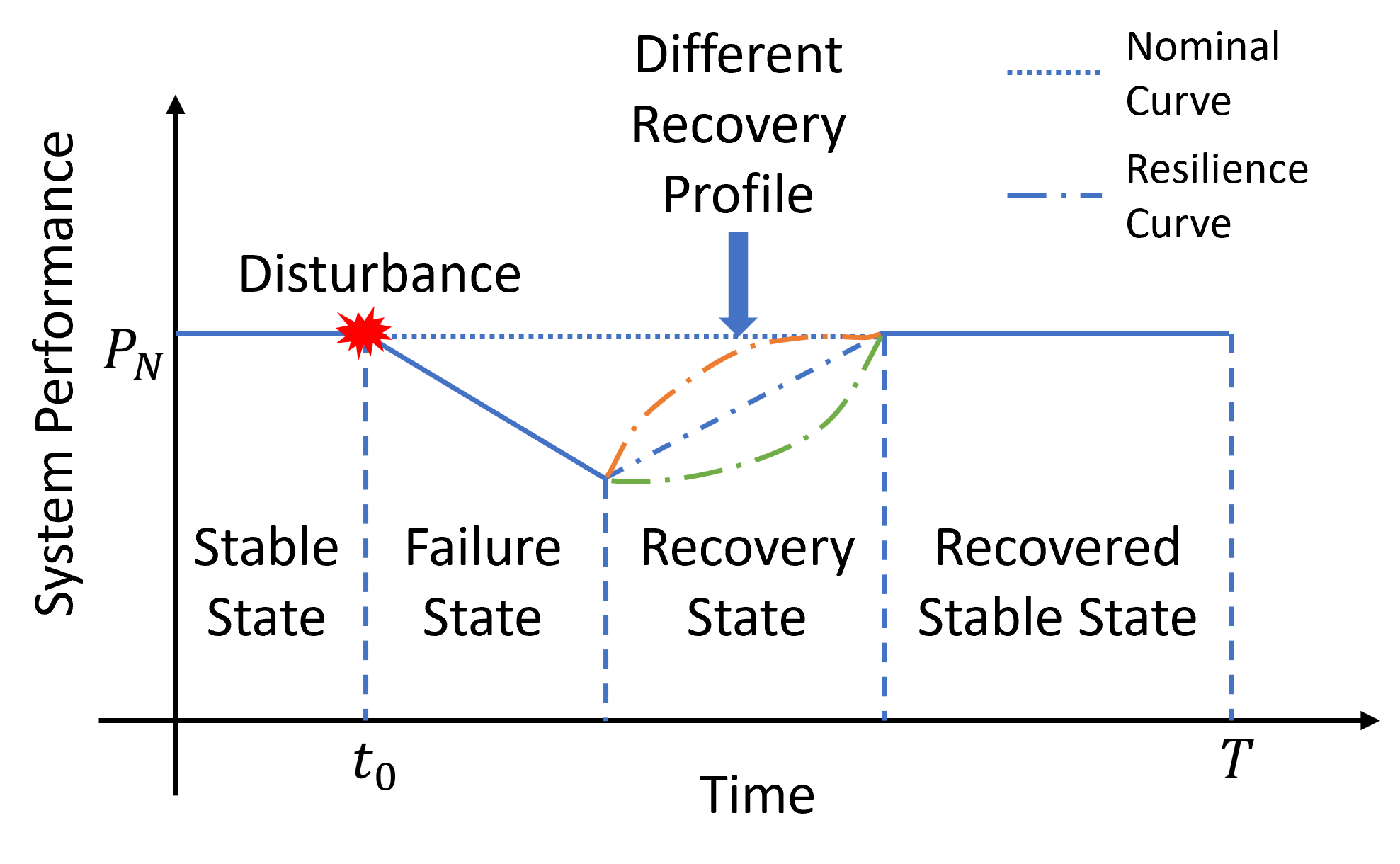}
    \caption{The resilience curve and four states in an ICI after with the disruptive event}
    \label{fig:resilience_curve}
\end{figure}

On the other hand, to measure the system resilience, a typical resilience curve (real performance curve after disruptions) with four states is illustrated in \autoref{fig:resilience_curve}. Note that the system performance curves could be different due to different strategies during the recovery process.
Based on the resilience curve, this study defines the resilience level with respect to changes in the system performance after the disturbance. It can be measured by comparing the resilience curve with the nominal system performance curve. In other words, the resilience metric is derived from the ratio of the area under the resilience curve to the area under the nominal performance curve. The specific formulation is:
\begin{equation}\label{eqn:resilience}
    \Phi=\frac{\int_{t_0}^{T}C_R(t)dt}{\int_{t_0}^{T}C_N(t)dt},
\end{equation}
where $\Phi$ denotes the resilience level, $C_R$/$C_N$ are the resilience curve and nominal performance curve respectively. $t_0$ is the initial time before the occurrence of the disruption, while $T$ is the termination time of the recovery process and the system settles at a new stable state. Intuitively this is true since a larger area under the resilience curve generally means a smaller performance loss induced by the disruption, and thus the system is more resilient considering a given disruptive event.

However, the above resilience term can only be measured after simulating a disruption event during the system nominal operation. This type of post-disruption information is hardly accessible during the design stage. Thus in order to bias the generative design process towards a more resilient system, a proxy is needed to represent the resilience level during the design stage. \autoref{sec:result} discusses the selection for the proxy in detail. And the actual resilience levels of the candidate designs become accessible through more expensive simulations during the post-processing stage.

\section{Modeling for Design Generator}\label{sec:generator}
Similar to discriminative methods, generative models have been well established for structured data, for example, the mixture model, the variational autoencoder, and the generative-adversarial network (GAN) \cite{bailey1994fitting, kingma2013auto, goodfellow2014generative}. Different from the traditional discriminative approach, the generative model tries to learn the underlying representation of the training data and tune the parameters of the model to create realistic outputs that are similar to the input structure. And this rationale enables the research of generative methods in the design community. Various research has been proposed to use the deep learning model for generative designs \cite{heyrani2021range, qian2022adaptive, oddiraju2022inverse}.
But it was not until recently that the generative models have been extended to unstructured data inputs i.e. graphs. In this study, we adopt the GVAE model, in which high dimensional representations of the nodes' information are encoded as latent vectors to further reconstruct the original graph. Notice that the focus of this study is on ICIs, and their nodes/edges have physical information representing the operational condition. That information is unique and crucial for reconstructing a physical system.

\subsection{Modeling}
%%%%%%%
%this section needs some improvement: when introduce the formulations, consider add the dimension of each function mapping, and vectors
%%%%%%%
As a VAE model, the GVAE algorithm contains two consecutive steps-the encoding and decoding. Thus following sections discuss the modeling of such an algorithm from those two aspects as well as how to train the corresponding data-driven design model.
 
\subsubsection{Encoding}
Similar to the standard VAE model, the GVAE first needs to encode the high dimensional node features $\mathbf{X}_v\in\mathbb{R}^f$ into latent vectors $z_v\in\mathbb{R}^d$, where $f$ denotes the number of features. And we use the Gated graph neural network (GGNN) model as the nodes embedder \cite{li2015gated}. The advantage of using the GGNN to embed the nodal information is that it can aggregate the neighborhood information in close proximity and preserve the information from the unique structure of each graph. And with the GGNN model, the input $\mathbf{X}_v$ is mapped into a multivariate diagonal Gaussian distribution in $d$-dimensional latent space, which are parameterized by $\mu_v$ and $\sigma_v$. And the latent representation of each node $z_v$ can be sampled from such a distribution. Following the convention of the standard VAE model, the regularization for encoding is the KL divergence between the latent distribution and the standard Gaussian distribution. Thus the loss term for the encoding step can be formulated as:
\begin{equation}
    \mathcal{L}_{encode}=\sum_{v\in G}KL(\mathcal{N}(\mu_v,diag(\sigma_v)^2)||\mathcal{N}(0,\mathbf{I}))
\end{equation}

\subsubsection{Decoding}
Different from the encoding part where the original graph information is embedded into latent vectors, the decoding aims to reconstruct a graph that is similar to the input data based on the encoded $z_v$. As for standard VAE with structured data inputs, the trained neural network for the decoder can automatically reconstruct the output through forwarding propagation. However, to reconstruct a graph, it's unclear how to directly output the whole graph in one single forward propagation process. Several studies have proposed to generate graph structures in an auto-regressive manner: starting from one node $i$, connect $i$ to nodes $j,k,m,...,$ which have the highest probability for connection, then keep growing the graph node by node until termination \cite{Liu2018, Liao2019}. Such an auto-regressive process involves two main decisions at each generation step: which node to connect, and use what type of edge to connect the new node. To answer those two questions, the GVAE model used as the generator takes four steps: node initialization, edge selection, edge labeling, and node updating. And we discuss the formulations of these steps as follows.

\begin{figure}
    \centering
     \includegraphics[width = \linewidth]{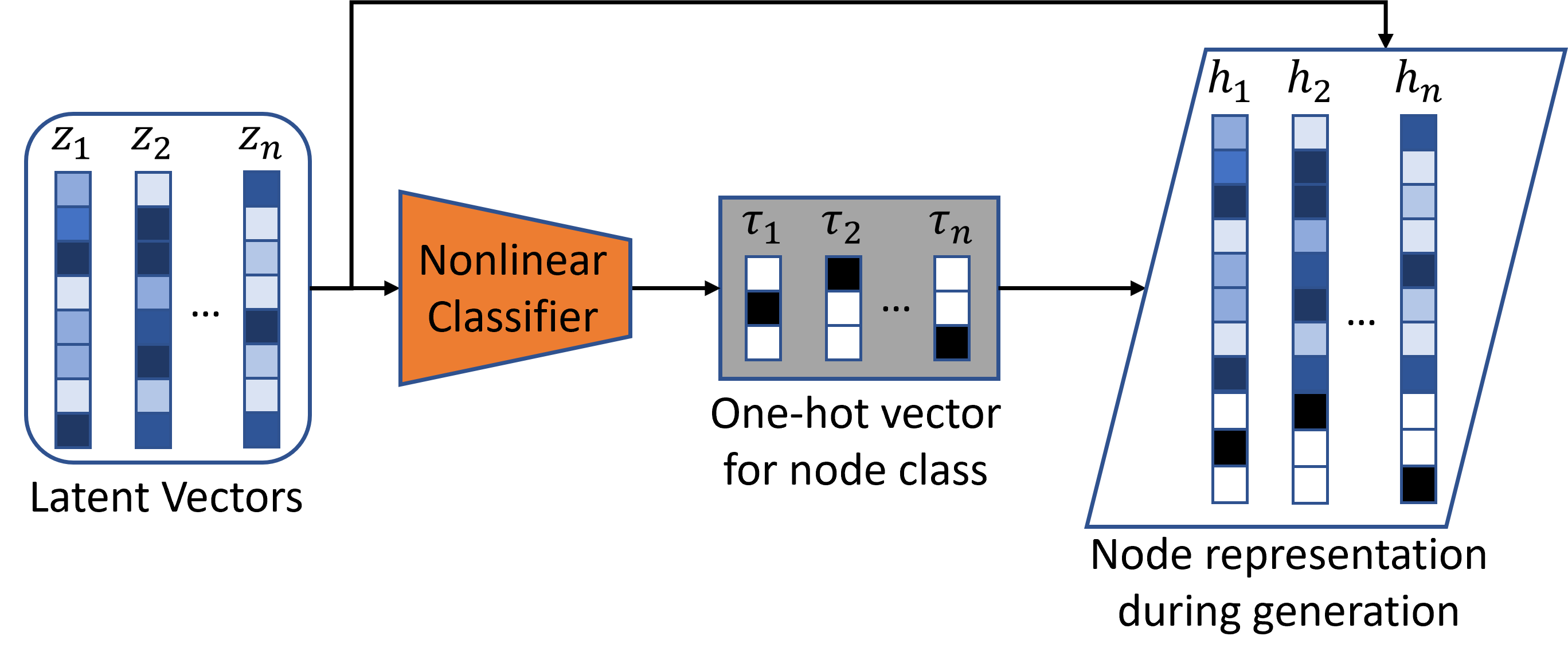}
    \caption{Node initialization for the decoding process: the latent vectors $z_v$ has been concatenated with their predicted nodal type, denoting as one-hot vectors.}
    \label{fig:generator_node_init}
\end{figure}

During the node generation process, it's crucial to determine which information to be utilized for initializing the node representation. Since the encoding process has already preserved the physical features of each node by mapping them into a latent distribution, only the label information remains to be taken care of. And for physical systems, it's important to consider the type of each node during the reconstruction process. For example, the nodes inside a power system could have three different classes-the generations, the load bus, as well as transmissible node. And the final generated design cannot randomly arrange disparate nodes violating physical constraints: they must follow the same pattern learned from the training dataset. Thus the node representation $h_v$ during the encoding step consists of two parts--the initial latent vector $z_v$ is concatenated with the one-hot class vector $\tau_v\in \mathbb{R}^{c}$ as shown in \autoref{fig:generator_node_init}. $\tau_v=f(z_v)$ where $f(\cdot)$ could be any appropriate nonlinear classifier for deducing the node labels. And in this study, we use a three-layer fully connected neural network as the $f(\cdot)$.

Once each candidate node of the graphs has been initialized, the graph generation process can start by establishing edges between nodes. Considering the current starting node is $v_i$, the task is to derive the probability of connecting $v_i$ to all other candidate nodes $v_j$ by using an edge $ij_l$ (selection), where $\ell$ is the type of the connection (labeling). Notice that unlike traditional generative algorithms on graphs, where the edges are indifferent, here the edges in ICIs carry rich physical information. For example, in a power grid, the decision-maker should assign different capacities to edges according to the magnitude of power loads at both ends. And to accomplish the tasks for edge selection and labeling, the feature representation of the candidate edge can first be constructed by concatenating various vectors:
\begin{equation}\label{eqn:edge_representation}
    \phi_{i,j}^{(t)}: = [h_i^{(t)},h_j^{(t)},{d_{i,j}},{H_{init}},{H^{(t)}}],
\end{equation}
where $h_i$, $h_j$ are the nodal representations; $d_{i,j}$ is the distance measure between node $i$ and $j$, for instance, the electrical resistance between two buses in a power grid; $H_{init}/H$ is the initial/updated global graph feature, which is defined as the average of all nodal representations in this study:
\begin{equation}\label{eqn:graph_representation}
    H^{(t)}:=\frac{1}{n}\sum_{i=1}^n {h_i^{(t)}}.
\end{equation}
\begin{figure}
    \centering
     \includegraphics[width = \linewidth]{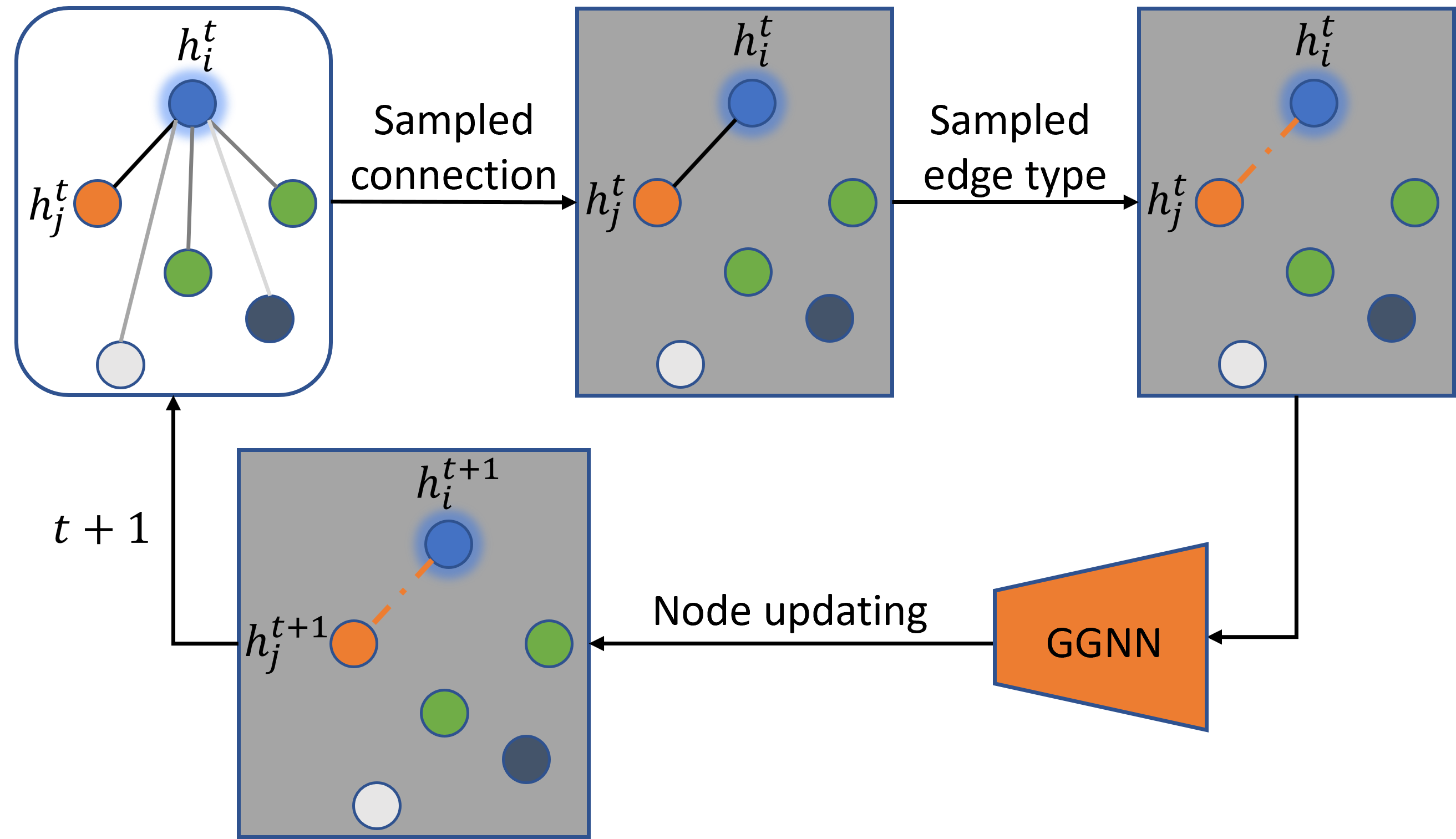}
    \caption{Illustration for the edge selection, edge labeling and the node updating process}
    \label{fig:generator_edge_update}
\end{figure}

Notice that all vectors in \autoref{eqn:edge_representation} and \autoref{eqn:graph_representation} have an extra index for time step $t$. This is due to the node updating procedure, which is discussed later. \autoref{eqn:edge_representation} shows the advantage of the auto-regressive generation process since the generation result not only depends on local information e.g. $h_v$, but also considers the global state of the graph $H^{(t)}$. Once the feature vector for candidate edge is established, we can start to model the probability of connecting edge $ij$ using type $\ell$ connection. First, given the edge feature vector $\phi_{i,j}^{(t)}\in\mathbb{R}^f$, the distribution of choosing to connect $ij$ via edge type $\ell$ is expressed as the product of the probability of connecting $i$ to $j$ and the probability of using type $\ell$:
\begin{equation}
    P(i\xleftrightarrow{\ell}j|\phi_{i,j}^{(t)}) = P(\ell |\phi _{i,j}^{(t)}, i\leftrightarrow j)P(i \leftrightarrow j|\phi_{i,j}^{(t)}).
\end{equation}
These two probability terms can be further calculated by formulating softmax functions:
\begin{equation}\label{eqn:softmax_edge}
    P(i \leftrightarrow j|\phi _{i,j}^{(t)}) = \frac{{M_{i \leftrightarrow j}^{(t)}\exp [C(\phi_{i,j}^{(t)})]}}{{\sum\limits_w {M_{i \leftrightarrow w}^{(t)}\exp [C(\phi _{i,w}^{(t)})]} }},
\end{equation}
\begin{equation}\label{eqn:softmax_edgetype}
    P(\ell |\phi_{i,j}^{(t)}) = \frac{{m_{i\xleftrightarrow{\ell}j}^{(t)}\exp [{L_\ell }(\phi _{i,j}^{(t)})]}}{{\sum_k {m_{i\xleftrightarrow{k}j}^{(t)}\exp [{L_k}(\phi _{i,j}^{(t)})]} }}.
\end{equation}
In the above two probability terms, $C$ and $L_\ell$ represent two trainable, nonlinear functions, e.g. fully connected neural networks, that map the feature vector to a scalar score. Moreover, additional masking matrices $M_{i \leftrightarrow j}$ and $m_{i\xleftrightarrow{\ell}j}$ enforce any prior knowledge about the generated graph. For example, node $i$ and $j$ cannot be connected because they are from different regions, or node $i$ and $j$ can only be connected by a high capacity edge $\ell$. After obtaining the probability distribution of the edge connection and labeling, new connection $ij_\ell$ can be sampled from the empirical distribution and the graph grows sequentially.

Notice that after choosing to connect node $i$ to node $j$, the graph structure has been changed due to the newly introduced edge and node. This changing in graph structure leads to changes in the node representation $h_i^{(t)}$, since nodes in the proximity of node $i$ are shifting. To update the node representation, GGNN can be used to re-derive $h_i^{(t)}$ in a recursive manner:
\begin{equation}
    h_i^{(t)}=GGNN(h_i^{(t-1)}, \sum_{j|ij\in E}h_j^{(t-1)}),
\end{equation}
where the initial node feature vector is $h_i^{(0)}$, and the updating aggregates all the feature vectors of nodes that are adjacent to $i$ in the current candidate graph design. In summary, the aforementioned edge selection, labelling and node updating steps are illustrated in \autoref{fig:generator_edge_update}. And the corresponding decoding loss can be formulated as:
\begin{equation}
    \mathcal{L}_{decode}=\sum_{G\in{\mathcal{D}}} \log(P(G|G^{(0)})\cdot P(G^{(0)}|z_v)),
\end{equation}
where the loss term essentially measures the log-likelihood of reconstructing the graph $G$ given in the dataset $\mathcal{D}$, with the initial encoded latent vectors $z_v$.
Due to space limitations, we have omitted some detail of the decoding process, including the termination condition for the graph generation. Readers can refer to \cite{Liu2018} for more in-depth discussion.

\subsection{Performance Aware Generation}\label{sec:performance_training}
\begin{figure}
     \centering
     \begin{subfigure}[b]{\linewidth}
         \centering
         \includegraphics[width=\textwidth]{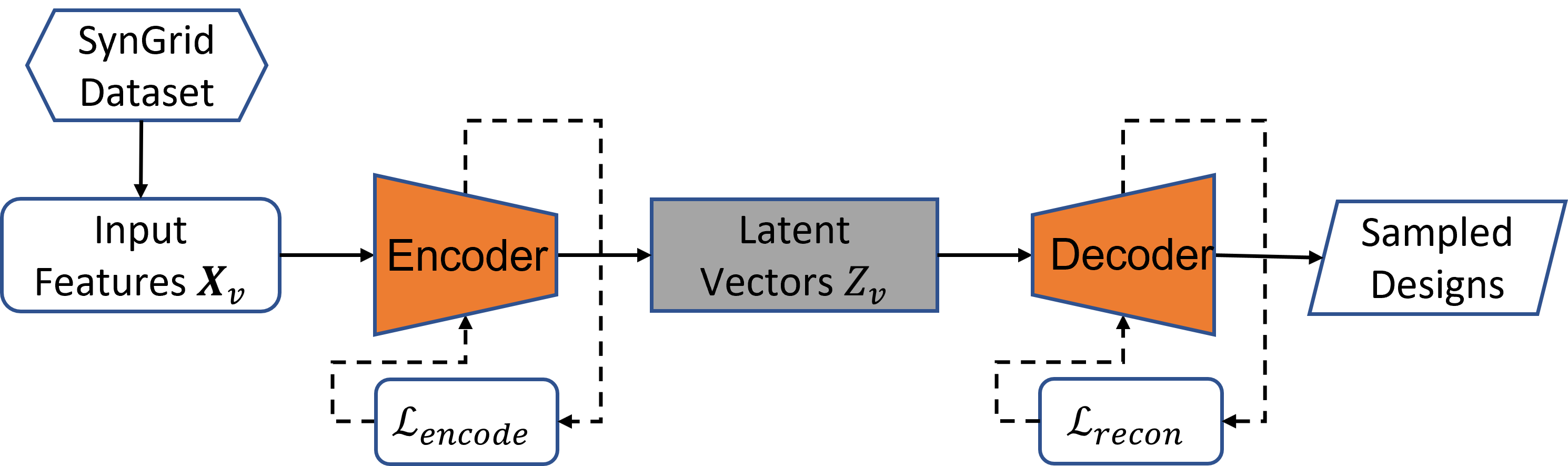}
         \caption{Training the GVAE model with the standard encoding and decoding loss terms}
         \label{fig:generator_encode_decode}
     \end{subfigure}
     \begin{subfigure}[b]{\linewidth}
         \centering
         \includegraphics[width=\textwidth]{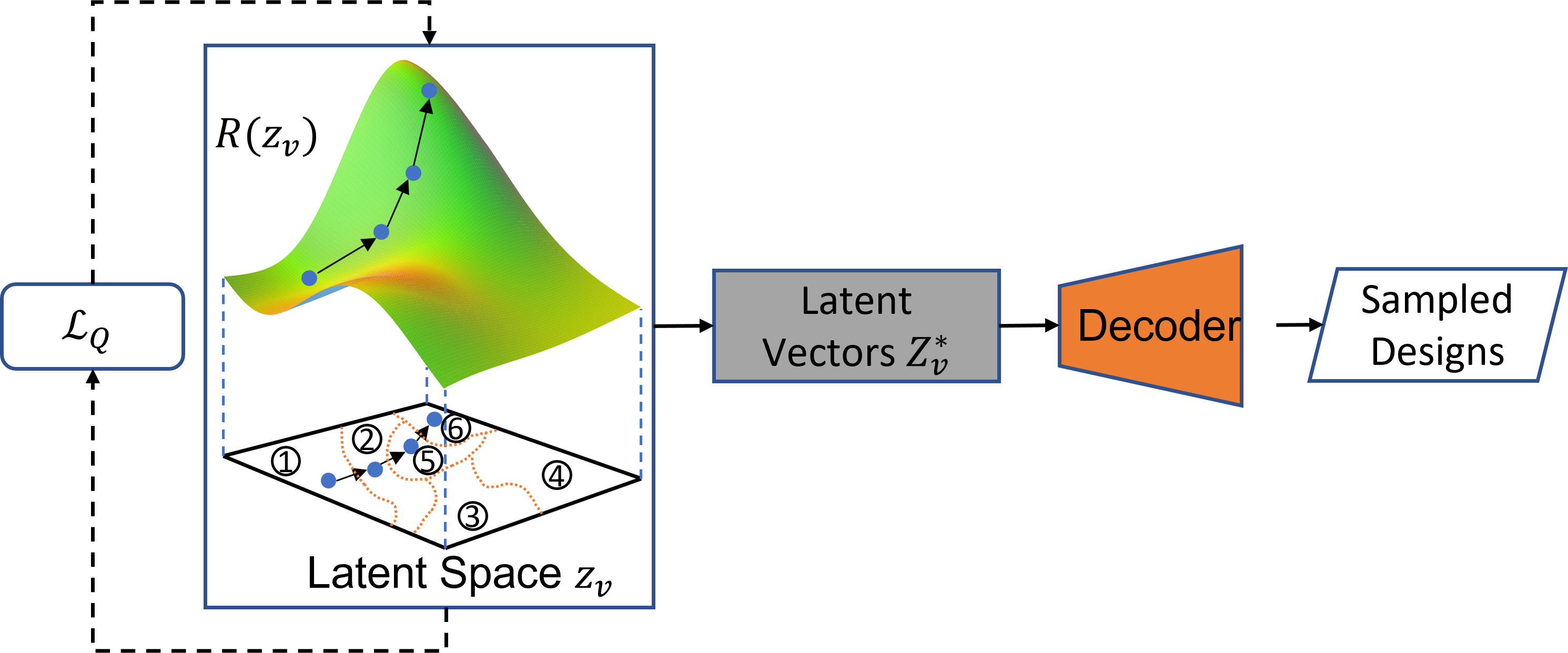}
         \caption{Performance aware generation by utilizing another surrogate $R(z_v)$ and gradient ascent wrt $z_v$}
         \label{fig:fig:generator_performance_aware}
     \end{subfigure}
        \caption{Flow chart for training the GVAE based design generator}
        \label{fig:figure5}
\end{figure}
So far, the basic loss terms for the encoding and decoding processes of the GVAE have been discussed. However, the conventional GVAE model with the $\mathcal{L}_{encode}$ and $\mathcal{L}_{decode}$ can only reconstruct candidate designs that are structurally similar to the samples in the input dataset. Since the established two loss terms only focus on quantifying the errors when encoding the latent design spaces and reconstructing from the encoding. To enable a generative design for resilience, the target performance criteria, i.e. the resilience level, needs to be correlated to the ICIs encoded in the structured, latent design space. As a result, another nonlinear mapping between the latent vectors and the performance criteria is constructed as:
\begin{equation}
    R(z_v)=\sum_{v} \sigma(f_1(z_v))\cdot f_2(z_v),
\end{equation}
where $f_1$ and $f_2$ are two fully connected neural networks. Notice that this formulation for estimating the performance based on latent vectors of the encoded ICIs adopts the idea from the GGNN regression model \cite{gomez2018automatic}. After training the $f_1$ and $f_2$ by using the training dataset, gradient ascent with respect to the input latent vector $z_v$ for $R(z_v)$ can be performed to find promising new candidate designs. And the loss term for finding the locally optimized $z_v^*$ is defined as:
\begin{equation}\label{eqn:performance_gen_loss}
    \mathcal{L}_{Q}=||R(z_v), Q||_2,
\end{equation}
where $Q$ is a predefined performance criterion. In other words, starting from a latent vector of the encoded ICI, the design with better performance can be found by following the direction within the latent space to increase the performance metric, as shown in \autoref{fig:figure5}. Thus, the final design can be decoded from the optimized $z_v^*$ by using the trained decoder model afterwards.

\section{Modeling for Design Estimator}\label{sec:estimator}
Although the performance training enables the generator to output desirable designs that optimize the predefined performance criteria, the candidate designs are still confined to the latent design space of the input dataset. Moreover, the generator only uses a simple fully connected neural network model to predict the performance for efficiency. To further bias the design towards optimal design space and to fine-pick the "good" candidates, a more sophisticated performance estimator is needed. As a design estimator, it needs to learn the mapping between the input information (raw information of the ICI) and the output scalar, i.e., the performance metric. And deep learning framework is well known to have superb performance for predicting numerical labels from high dimensional training data. As a result, in this study, we adopt the graph convolutional network (GCN) \cite{Hammond2011, Defferrard2016, Kipf2017} to construct the performance estimator for the graph-like ICIs.

\subsection{Formulation}
The convolutional neural network (CNN) is well known for its excellent performance on image classification for nearly two decades~\cite{LeCun1998}. However, the application of CNN on 2D Euclidean space is not extended to unstructured topological space until Bruna et al.~\cite{Bruna2014} propose the generalization of CNN to signals defined on the graphical domains. This extension significantly improves the applicability of the convolution process on unstructured data, such as information generated from social networks or chemical compounds. After their work, the research about convolution on graphs diverges into two directions: one is working on the spatial domain of graphs while the other integrates the spectral graph theory into the convolution process. And the breakthrough works in \cite{Hammond2011, Defferrard2016} have greatly increased the potential of the spectral GCN, by proposing an efficient first-order approximation of spectral graph convolutions. This study utilizes the latter approach, i.e., the spectral GCN. And the GCN takes the advantage of the underlying information embedded in the adjacency matrix to perform the learning tasks for networks. Therefore GCN has superb performance for the inference task on graph inputs.

It has been proved that the GCN can be derived from the traditional CNN by studying the spectral graph theory \cite{Hammond2011, Defferrard2016}. GCN can be treated as the generalized case of the CNN on arbitrary, unstructured space. And following the formulation in \cite{Kipf2017}, the propagation rule, or the convolutional operator for graphs requires the degree and weighted adjacency matrix as additional parameters. Different from conventional adjacency matrix with binary entries only, the weighted one can include other values than 0/1 representing the edge weights between different pairs of nodes. This characteristic enables the learning task fulfilled by GCN to take the advantage of the rich graphical information of ICIs. On the other hand, the degree matrix $D$ is diagonal and aggregates the neighborhood information for each node. It can be obtained based on $A$: $D_{ii}=\sum_{j}A_{ij}$. With the degree and adjacency matrix on hand, the GCN propagation rule is defined as:
\begin{equation}\label{eqn:graph_conv}
    GCN(X)=D^{-1/2}AD^{-1/2}X\Theta,
\end{equation}
where $\Theta$ are the trainable weights of each convolutional layer. Therefore, the graph inference problem can be solved by constructing a deep neural network (DNN), for instance $q=\sigma(f_2(f_1(GCN_2(GCN_1(X)))))$ where $GCN_1/GCN_2$ are the convolutional layers defined in \autoref{eqn:graph_conv}, and  $f_1/f_2$ are adequate nonlinear functions.

\subsection{Training}
\begin{figure}
    \centering
     \includegraphics[width = \linewidth]{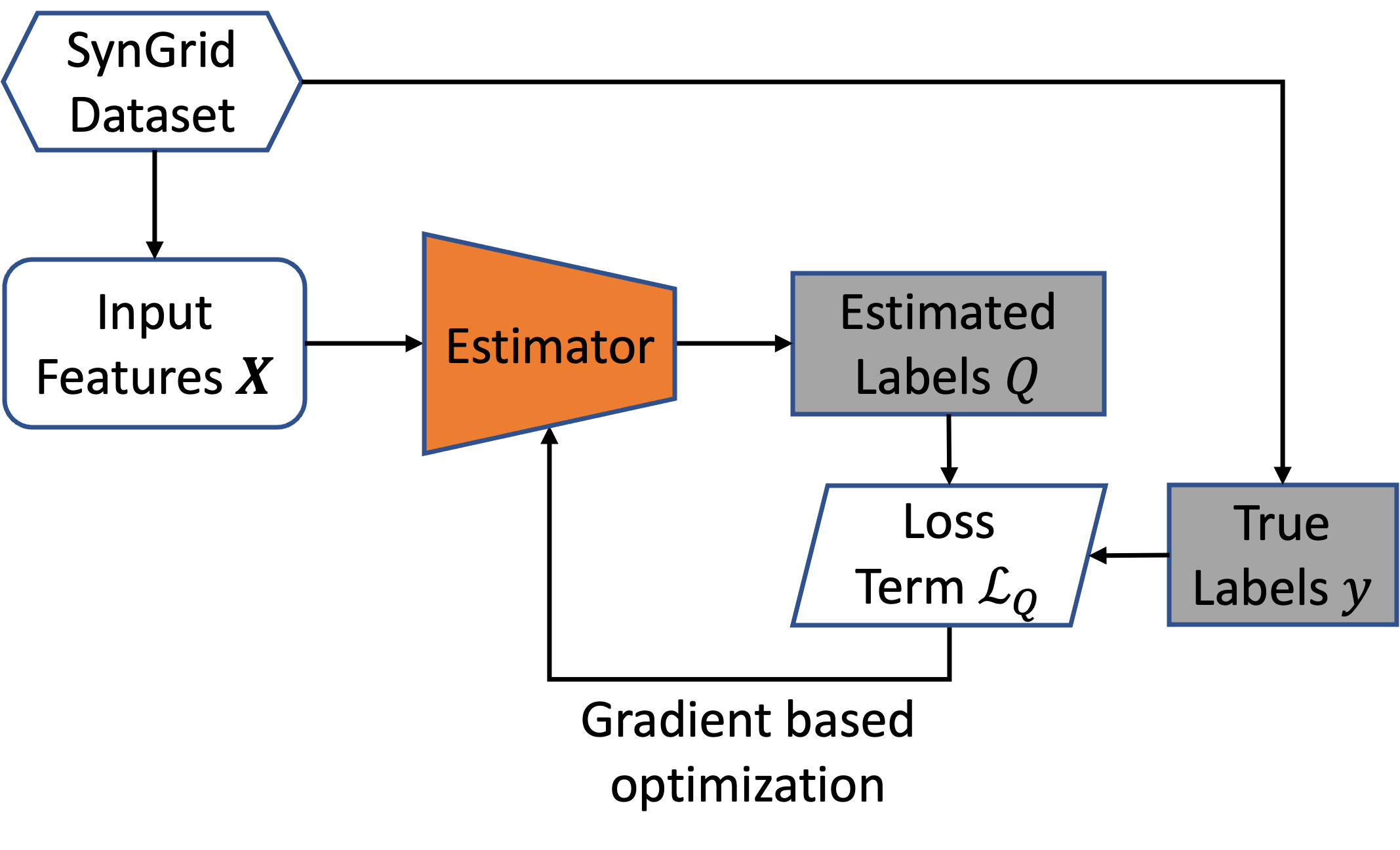}
    \caption{Flow chart for training the GCN based design estimator}
    \label{fig:estimator_flowchart}
\end{figure}
As for using the GCN algorithm to evaluate the performance of generated designs, the specific training process needs to be discussed. Like conventional DNNs, the training process for the GCN-based estimator requires gradient information to optimize the trainable parameters as shown in \autoref{fig:estimator_flowchart}. The dataset of ICIs contains the input feature $\mathbf{X}\in\mathbb{R}^{N\times f}$, where $N$ is the number of nodes inside the graph and $f$ is the number of features considered. For instance, for the 123-bus power grid shown in \autoref{fig:sample_network}, $\mathbf{X}$ contains $123$ rows and $6$ columns, that includes the information of the power demand magnitude, generation capacity level, the voltage angle, etc. Whereas the label $y$ used for training the estimator is a numerical value for each training sample. Since the generation process needs to be biased to have a more resilient system design, the label should represent the overall system resilience level. And the choice for the label $y$ is further discussed in \autoref{sec:result}. The loss term used to train the estimator is the mean squared error (MSE) between the estimated label $q$ and the actual performance $Q$. The formulation is similar to \autoref{eqn:performance_gen_loss}. Notice that, unlike the design generator, the estimator can be pretrained on the existing training dataset so that estimating the performance of new designs is computationally cheap during the iterative generating process.

After training a GCN-based design estimator, we can combine the module with the generator to form an iterative process for refining the designs. As shown in \autoref{fig:main_flowchart}, a fixed number of $z_v$ is sampled and are used to generate the corresponding designs. Then the estimator determines the best batch of designs in terms of metric $Q$. If the process is not converged, i.e. the difference between the last best design and the current optimal one is large, then the top $c$ generated designs are mixed with the original dataset to perform another round of generation. To improve gradually the generated samples, the batch of the best designs replace the group of the original training data that has 

\section{Post-processing Simulation}\label{sec:postprocessing}
So far the performance criterion $Q$ for training the design generator/estimator has been discussed. And for an efficient training process, the performance of each design is represented by an easy-to-evaluate scalar metric: rather than the actual resilience level, it is a proxy to access the system resilient performance. Recall the system resilience mainly measures the system's capability toward unforeseen disruptions. And it's hard to obtain the resilience level without running experiments for systems undergoing disturbances. As a result, we add a post-processing stage to further evaluate the candidate designs filtered by the surrogate metric $Q$. Since the initial design space has been shrunk from millions of candidates to hundreds through the iterative biasing generation, it becomes tractable for running more complex, simulation based evaluations during the post-processing stage.

Following the convention of the studies for ICI, here the metric used in the post-processing simulation is the expected demand not supplied (EDNS). And it's defined as:
\begin{equation}\label{eqn:EDNS}
    EDNS = \sum_{e_i\in S_{e_i}} P_{e_i}C_{e_i},
\end{equation}
where $e_i$ denotes a possible disruptive event and $S_{e_i}$ is the set of all events simulated. $P_{e_i}$ represents the probability of having the event $e_i$, and $C_{e_i}$ is the amount of lost demands after $e_i$. In \autoref{fig:resilience_curve}, the EDNS can be viewed as the area between $C_R$ and $C_N$, which quantifies the capability of the system withholding external disruptions. And the optimal design can be determined as the one with the smallest EDNS after acquiring the post-processing results.

\begin{figure}
    \centering
    \includegraphics[width = \linewidth]{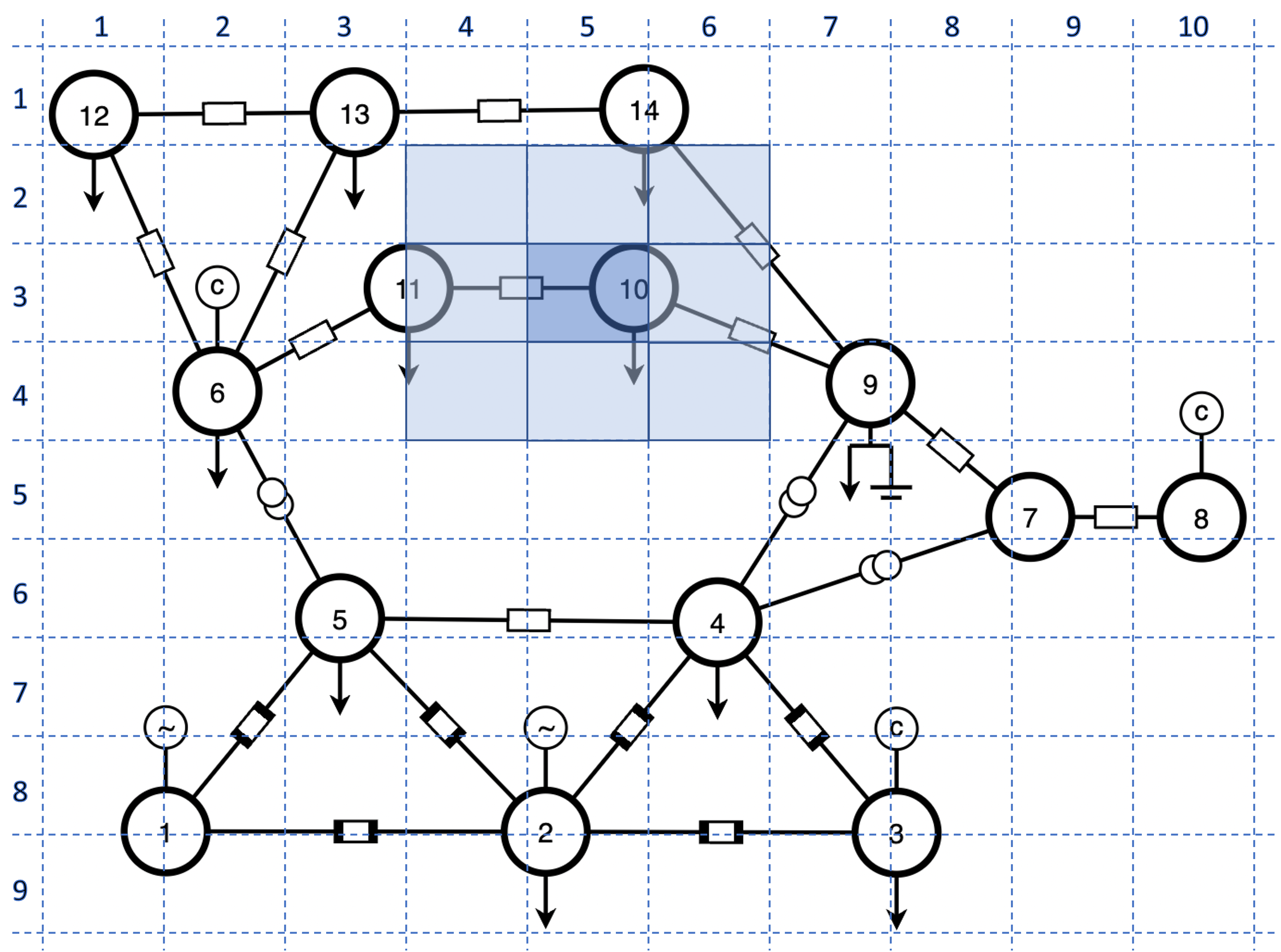}
    \caption{A small scale power system mapped to an example meshgrid with a disruptive event sampled at cell (3,5)}
    \label{fig:mesh_grid_postprocessing}
\end{figure}

To obtain the $C_{e_i}$ after each $e_i$, we establish meshgrids for mapping each candidate design with geological information. \autoref{fig:mesh_grid_postprocessing} illustrates the mesh view of the IEEE 14-bus system with a disruptive event. Notice that, the mapped mesh-view of the system is helpful for simulating realistic disruptive events. Since external disruptions are usually confined within a specific region. For example, as shown in \autoref{fig:mesh_grid_postprocessing}, a storm with predefined $P_{e_i}$ is simulated to happen at cell (3,5), which leads to the highest probability of failure for node 10 and edge (10,11). Also, the components within the close proximity of cell (3,5), including node 10, 14 and edges around, incur smaller failure probability. Thus based on the information from the meshgrid, the curtailed performance of the ICI caused by random disasters can be simulated. For instance, $C_{e_i}$ from solving the optimal power flow problem for power systems, or obtaining the maximum flow solution for an impacted supply chain network, after disconnecting the damaged components. With each $P_{e_i}$ and the corresponding $C_{e_i}$, the overall EDNS of the candidate design can be found by performing several runs of simulations in a Monte Carlo manner.

%%%%%%%
% provide relevant results about post-processing
%%%%%%%
%%%%%%%
%need to provide the exact numbers used in the generator/estimator step in Case study section
%%%%%%%
%%%%%%%
%add a small test case for supply chain network design
%%%%%%%
\section{CASE STUDIES}\label{sec:result}
To validate the proposed generative design method, experiments about designing different kinds of ICIs are considered. \autoref{sec:synthetic} presents the design results for small-scale synthetic network systems. And \autoref{sec:power_system} further discuss the applicability of the proposed framework on large-scale IEEE test feeders.

\subsection{Synthetic Network Systems}\label{sec:synthetic}
\begin{figure}
     \centering
     \begin{subfigure}[b]{0.49\linewidth}
         \centering
         \includegraphics[width=0.85\textwidth]{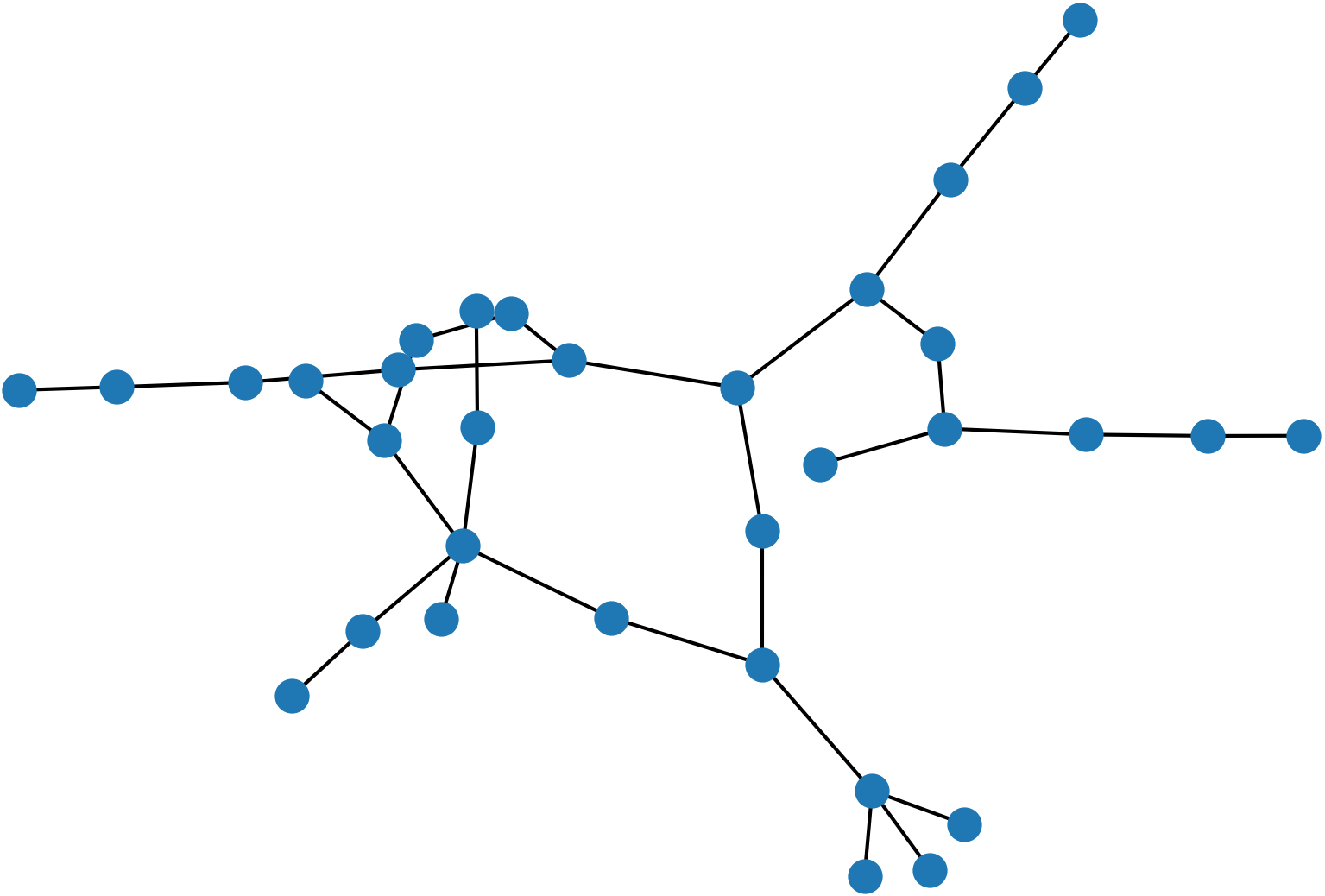}
         \caption{}
         \label{fig:syn_1}
     \end{subfigure}
     \begin{subfigure}[b]{0.49\linewidth}
         \centering
         \includegraphics[width=0.85\textwidth]{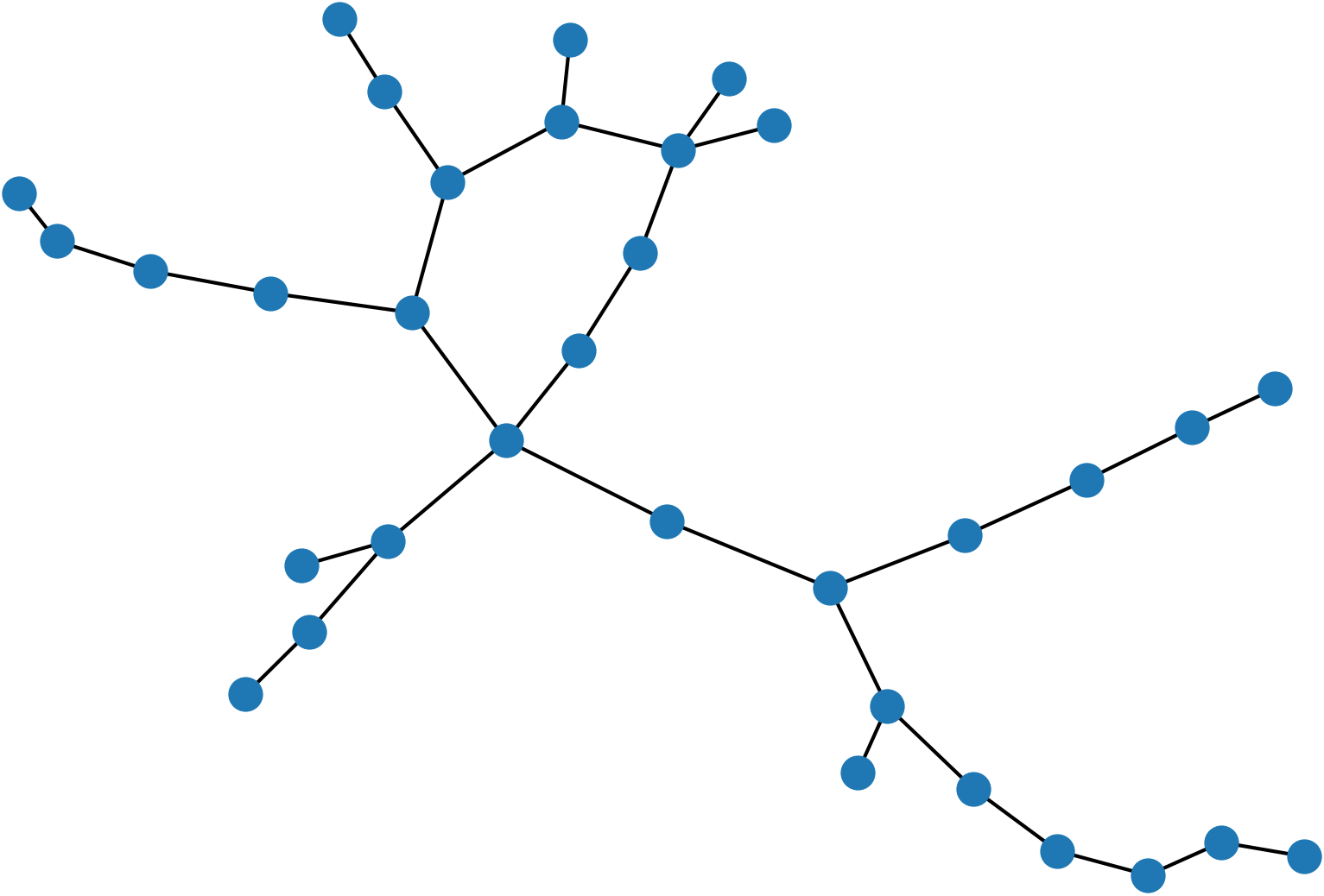}
         \caption{}
         \label{fig:fig:syn_2}
     \end{subfigure}
     \begin{subfigure}[b]{0.49\linewidth}
         \centering
         \includegraphics[width=0.85\textwidth]{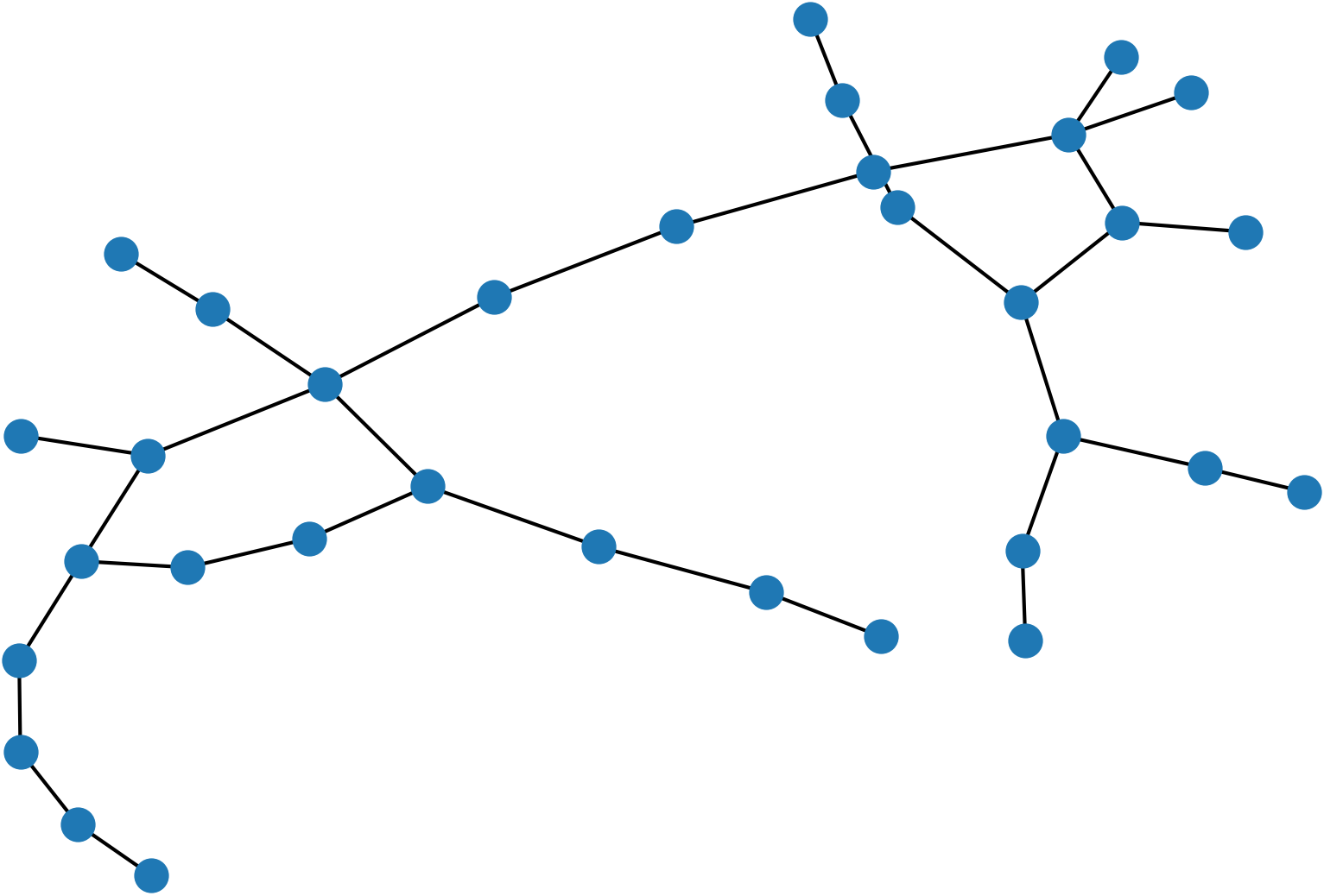}
         \caption{}
         \label{fig:fig:syn_3}
     \end{subfigure}
     \begin{subfigure}[b]{0.49\linewidth}
         \centering
         \includegraphics[width=0.85\textwidth]{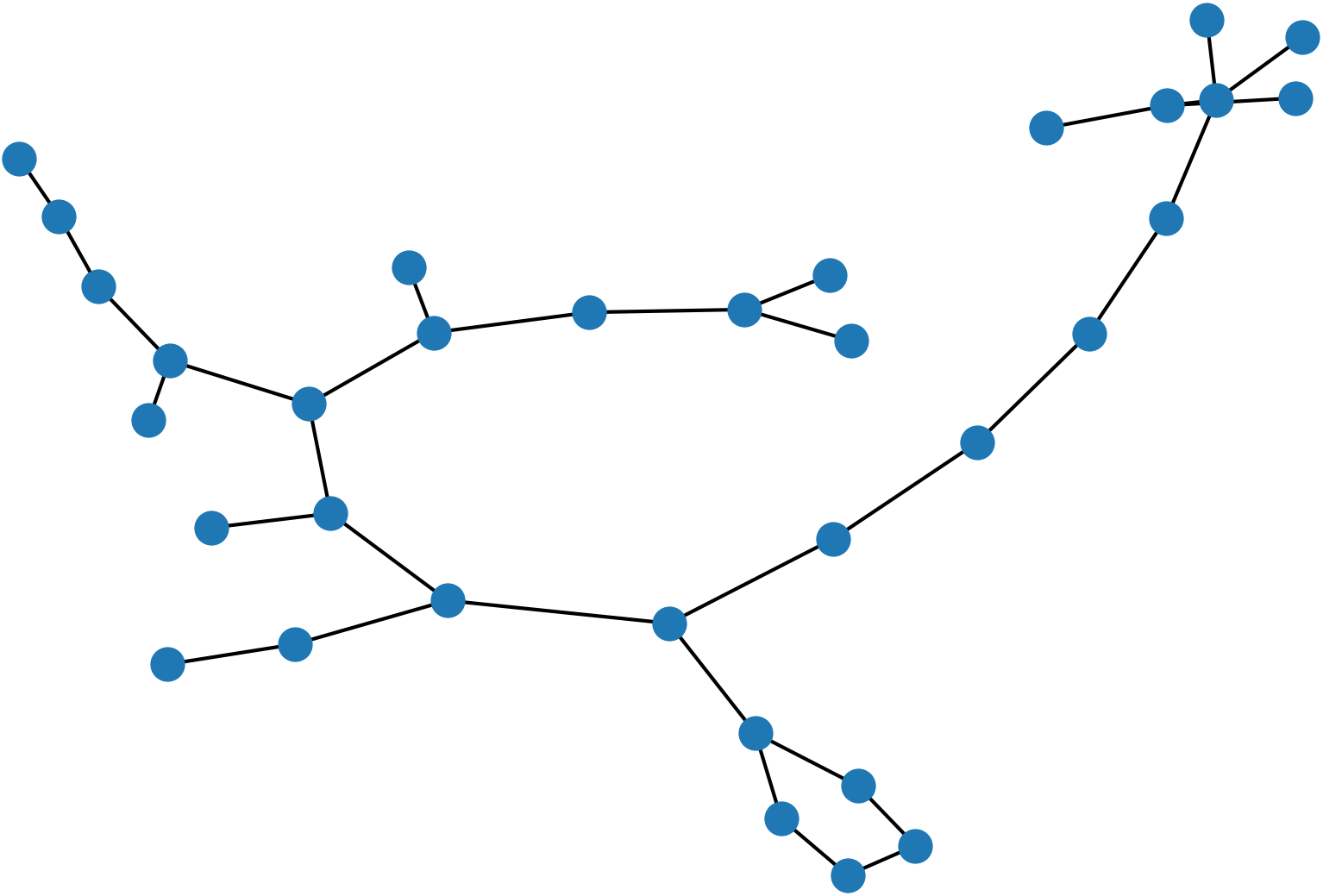}
         \caption{}
         \label{fig:fig:syn_4}
     \end{subfigure}
        \caption{Example small-scale Watts–Strogatz graphs used as the training data}
        \label{fig:sample_syn}
\end{figure}
First, we utilize the NetworkX package \cite{SciPyProceedings_11} in Python to randomly generate 10,000 Watts–Strogatz small-world graphs as the design training dataset. This type of random graphs has the small-world properties, which include high clustering and short average path length. Those properties enable researchers to use the Watts-Strogatz model to understand realistic networked systems, for instance, social networks and supply chain systems. \autoref{fig:sample_syn} demonstrates example Watts-Strogatz graphs generated for the case study. Here a small-scale system is considered, which includes 33 nodes and the same number of edges. We can see from \autoref{fig:sample_syn} that the initial graphs used for training has few clusters and several nodes have higher degree comparing to remaining nodes. This is desirable since for a supply chain network, nodes with higher degree can represent transportation hubs, while clusters indicate different coverages for separated communities.

\begin{figure}
    \centering
    \includegraphics[width=0.8\linewidth]{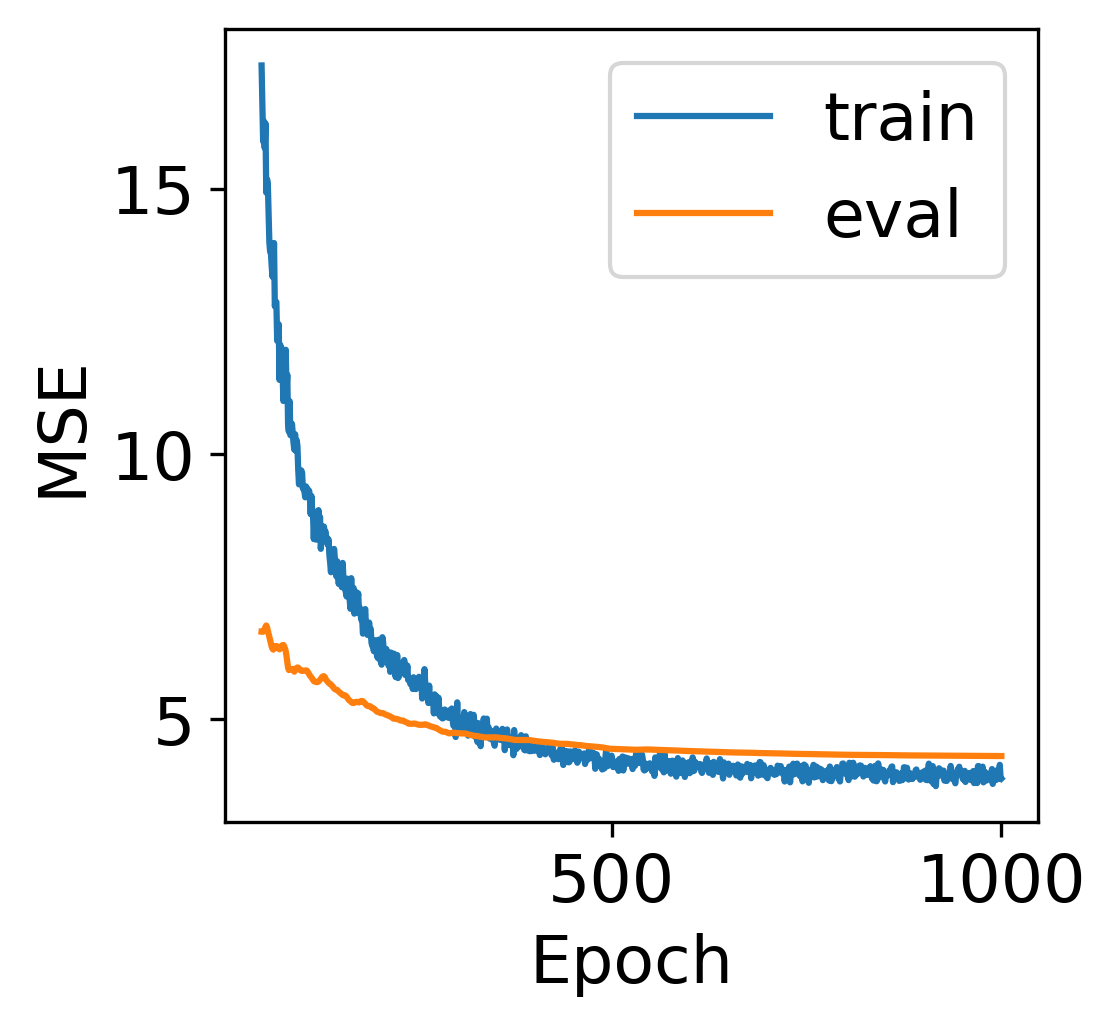}
    \caption{The training history for the estimator on synthetic network system data}
    \label{fig:estimator_loss}
\end{figure}

As for designing the ICI, we need to define the physical features of the nodes and edges. And in this case study the nodes have a numerical feature to indicate the magnitude of demands/supplies as well as a categorical feature declaring the type of the node: demand, supply and transfer. This categorical feature is pre-sampled based on the degree of each node. For example, a node with higher degree has much higher probability of being the supply node, while nodes with small degree are likely sampled as demand or transfer nodes. On the other hand, The edges have two numerical feature denoting the capacity and cost of each edge. Like classical network flow problems, the cost is measured in terms of per unit flow on the corresponding edge. And to quickly evaluate the design generated during the iterative process, the performance metric $Q$ is defined as the magnitude of maximum flow with the minimum cost inside the network, denoting as $f_{max}$. Notice that to derive the $f_{max}$, two dummy nodes are added to the sampled Watts–Strogatz graphs to represent the source and destination nodes. Once the training dataset is ready, the estimator can be pretrained for qucikly evaluate the designs during generation. \autoref{fig:estimator_loss} shows the training history of the GCN-based estimator, where the y-axis represents the MSE between the ground-truth and the estimated performance metric.

\begin{figure*}
    \centering
    \includegraphics[width = 0.8\linewidth]{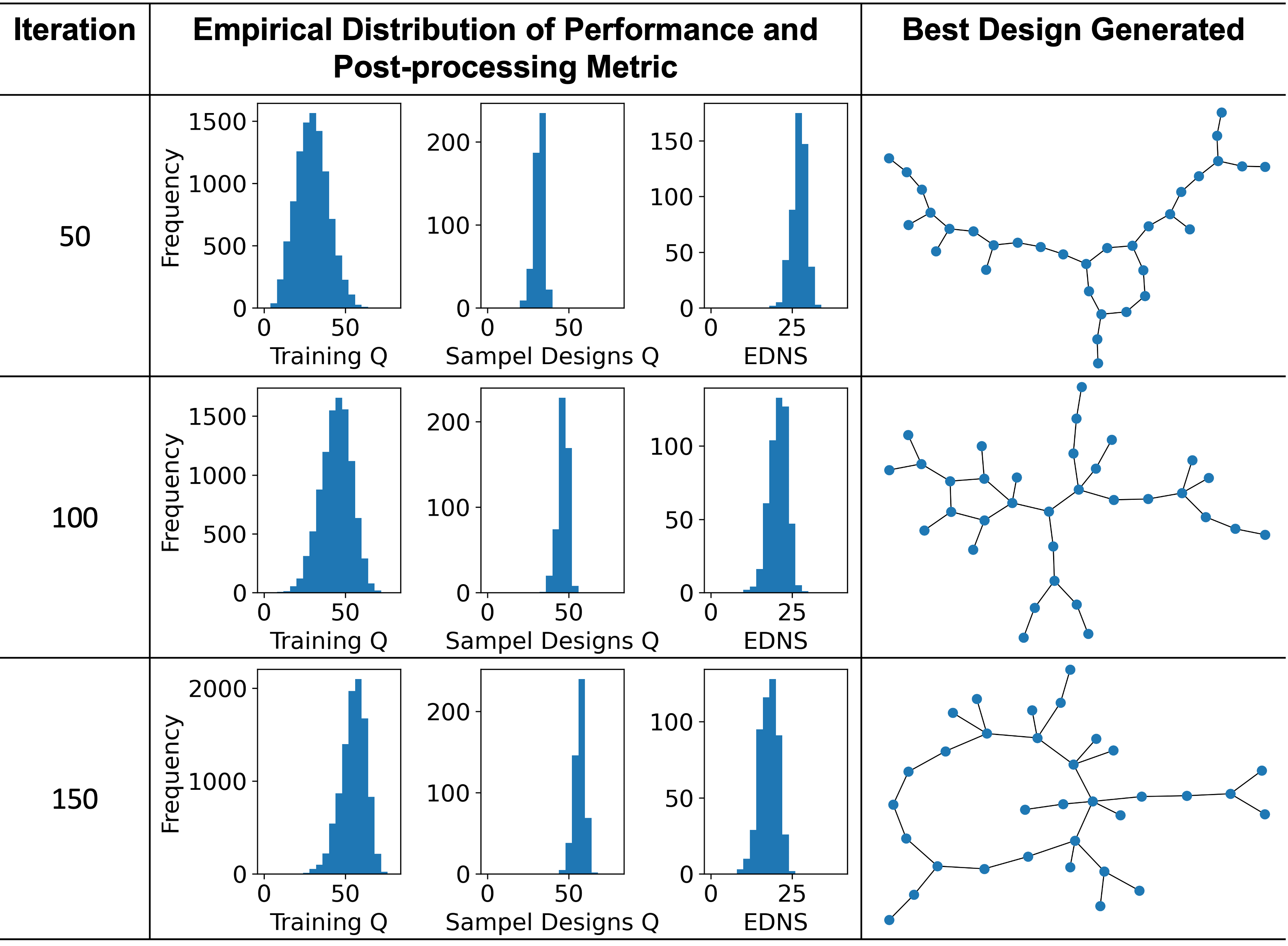}
    \caption{Snapshots of the empirical distribution of the metric Q of the training data and the best designs generated, as well as the post-processing EDNS metric at different training iterations}
    \label{fig:syn_result}
\end{figure*}

Based on the training dataset, our goal for the design is to come up with a new and optimal design that satisfies the demands within the network while maintains steady performance after disturbances. We set the termination criterion for the training process as reaching 150 iterations. \autoref{fig:syn_result} summarizes the distribution of the performance metric $Q$ of the training dataset and the 500 generated designs, as well as the best design evaluated by the estimator at each iteration. Moreover, to show the performance biasing design process for a resilient system, we also show the distribution of the EDNS of the 500 designs along with the generation metric $Q$. Notice that the in-production generative design process only needs to execute the post-processing step after the training has bee converged. Here to demonstrate the correlation between the training metric $Q$ and the EDNS, we conduct post-processing simulations even during the middle of the generation.

According to the results shown in \autoref{fig:syn_result}, the generative design method successfully biases the candidate designs toward predefined performance metric, especially for the training label $Q$. Though the changes in the resilience index EDNS are not significant at the same level as that of $Q$, there still exists a trend for improvement: the overall EDNS decreases significantly during the first 100 training iterations. On the other hand, take the best design generated at each iteration as an example, the first version of the design contains long branches and few redundant paths. This is not a good practice for designing a resilient ICI like supply chain network. After several iterations' training, the candidate designs become to have clustered communities as well as central hubs along with substations. Those hierarchical components lead to a much smaller EDNS. At the last iteration, the central skeleton of the network changes from a tree structure to a ring shape, which further introduces more redundancy to the already high clustering network. This shifting in design strategies also demonstrate the capability of the proposed framework on learning good proprieties from existing training samples.
\begin{figure*}
    \centering
    \includegraphics[width=0.8\linewidth]{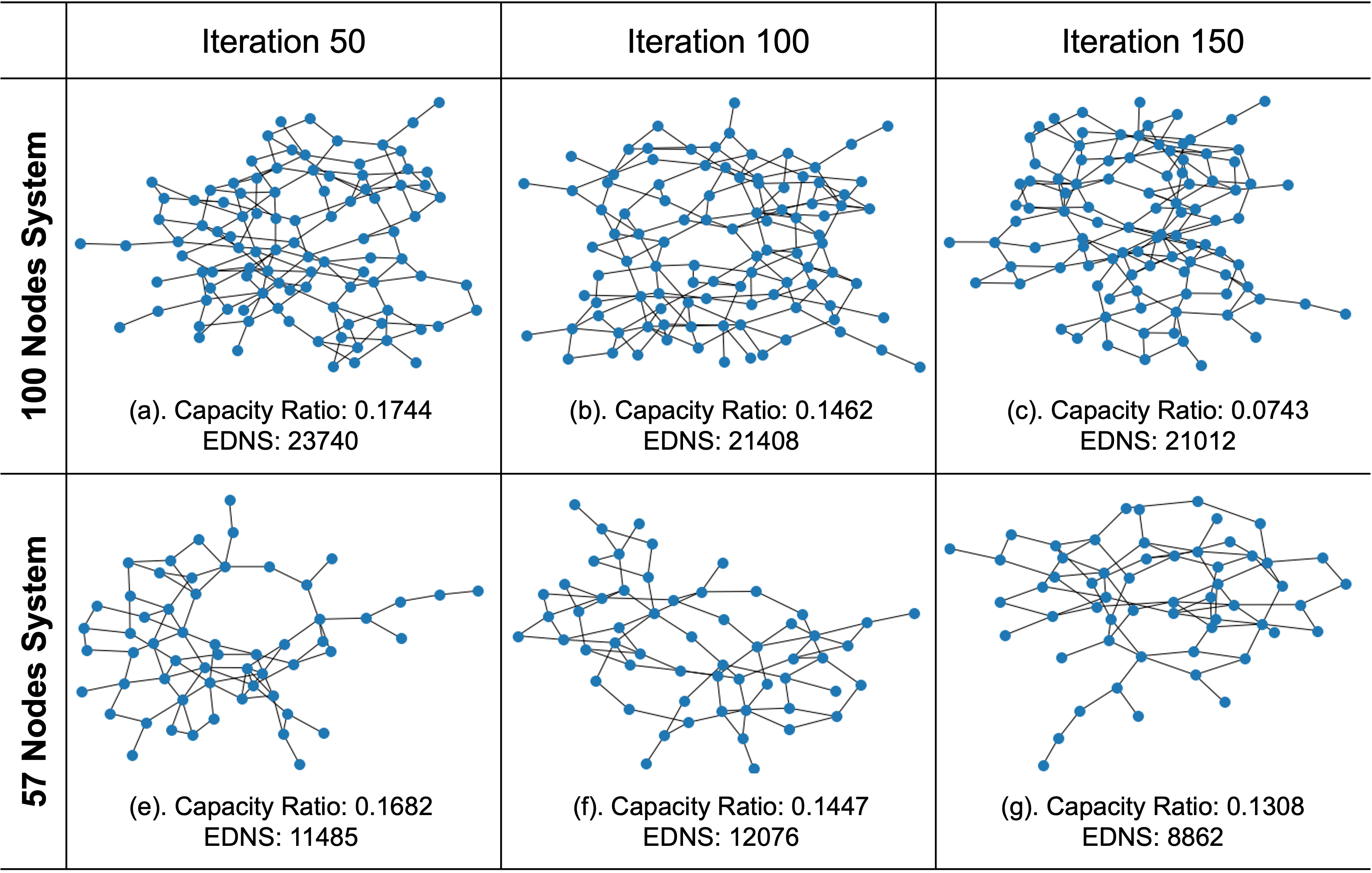}
    \caption{Illustrations of brand new designs generated as well as their capacity ratios and EDNS}
    \label{fig:brand_new_design_results}
\end{figure*}

\subsection{IEEE Power Grids Test Case}\label{sec:power_system}
\begin{figure*}
    \centering
    \includegraphics[width=0.8\linewidth]{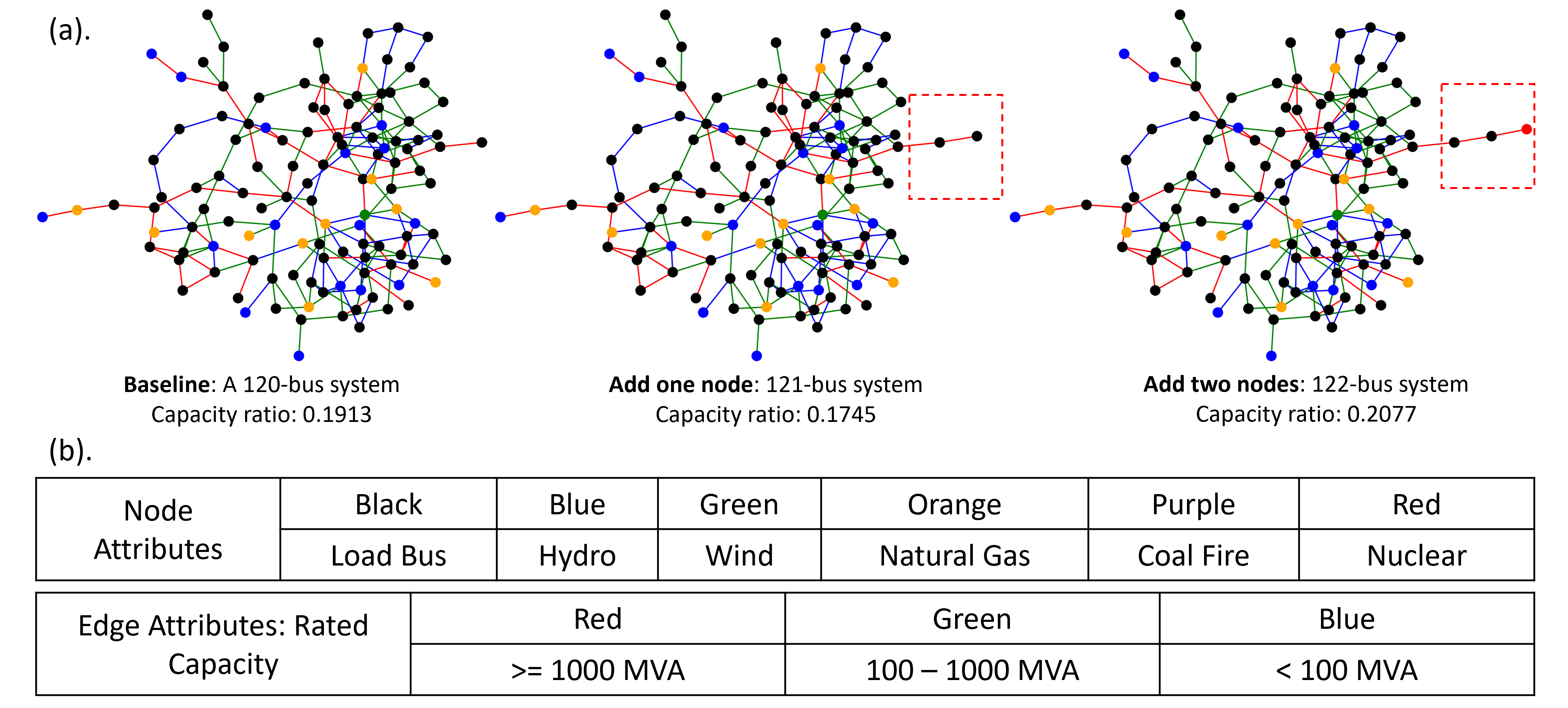}
    \caption{The solution of an expansion problem generated from the framework and the specific physical information considered}
    \label{fig:expansion_design_results}
\end{figure*}
Despite the small-scale test case based on synthetic networks, we also consider real world dataset about the power systems. In order to train the design estimator and the generator, we first generate a power grid dataset consisting of 138000 sample systems that range from 50 nodes to 150 nodes. The sample system designs are generated by the SynGrid package in MATPOWER \cite{wang2010generating}. And since the exact resilience level of a system design is hard to quantify without knowing the online response during disruptions, here we adopt a surrogate metric, the total capacity ratio of the system. This metric implicitly measures how well the system gonna behave towards external disruptions. And it is defined as:
\begin{equation}
    C^R = \frac{\sum_{ij|ij\in{E}}f_{ij}}{\sum_{ij|ij\in{E}}u_{ij}},
\end{equation}
where $f_{ij}$ is the actual flow on edge $ij$ from the optimal power flow (OPF) results; $u_{ij}$ is the capacity assigned to $ij$. A lower capacity ratio means that the design can have more room for unpredictable flows. And this metric can indicate the tolerance to failures for an ICI during disruptive events, e.g. sudden load surges or line outages. However, the final performance metric requires additional compensations to prevent the generated designs from assigning unnecessary large capacity to every edges for obtaining a low $C^R$. Therefore, we take the cost of having large edge capacity into consideration. And the total cost of establishing all edges with capacity $u_{ij}$ is denoted as:
\begin{equation}
    C^E=\sum_{ij|ij\in{E}}a\cdot u_{ij},
\end{equation}
where $a$ is a predefined parameter measuring the unit cost of edge capacity. The training label is then defined as:
\begin{equation}
    Q=\alpha C^R + (1-\alpha)\frac{C^E}{K},
\end{equation}
where $\alpha$ is the weight hyperparameter, and $K$ is a constant for normalizing. The training task is thus to minimizing $Q$. Notice that the proposed framework is not limited to any specific form of metric, and it can adopt other appropriate resilience-driven metrics by changing the formulation of the $Q$.

To show the applicability of the design framework, we trained the model on the SynGrid dataset and followed the steps depicted in \autoref{sec:performance_training} to generate brand new designs with a low level of capacity ratio. \autoref{fig:brand_new_design_results} shows the samples from the progress of the gradient ascent and the designs decoded from the embedded $z_v$. Here, a mid-scale 57 nodes system and a large-scale 100 nodes system are considered. Although the generative design process only uses a surrogate metric instead of the actual system resilience like EDNS, both the proxy metric used for training and the more expensive resilience index show the similar trend throughout the iterations. This finding suggests that the relatively cheap performance metric $Q$ is suitable for deriving a resilient design via the generative process. Moreover, different from designing a brand new system, the expansion problem sometimes is more prevalent For ICIs. For instance, where to add new buses in an existing power grid to enhance the current service, or how to allocate new transportation hubs in a supply chain network to improve the coverage. In order to solve an expansion problem, we can include the prior knowledge constraints i.e. existing connection information, as masking matrices $M_{i \leftrightarrow j}$ and $m_{i\xleftrightarrow{\ell}j}$ shown in \autoref{eqn:softmax_edge} and \autoref{eqn:softmax_edgetype}. Such an information injection during the generation process is also useful when the stakeholder requires to put hard constraints on the final design.

\autoref{fig:expansion_design_results} illustrates the expansion process for a 120-bus system after adding two additional nodes. Here the physical information in both the original ICI and the expanded system is included. Six types of nodes are considered and the edges have three categories based on their rated capacities. Notice that the generated designs from the GVAE model only contain categorical information for each node and edge, e.g. a node is assigned as a coal-fired power plant and an edge is allocated with high capacity. To conduct OPF simulations in the MATPOWER and to obtain the capacity ratio, we have sampled numerical ratings for generations and edges, based on the statistics in the SynGrid package.

\section{CONCLUSION}\label{sec:conclusion}
In this study, a generative design framework for networked systems has been proposed. Different from traditional model-based, optimization-driven design methods, this generative approach is generalized and requires no explicit model development. The framework utilizes advanced graph learning algorithms, including GCN and GVAE, to realize a model-free, data-driven design schematic. The proposed design framework is capable of mining intelligently useful properties from existing system designs and biasing the candidate designs to meet predefined performance criteria. Moreover, prior knowledge about the system can be included during the generation process to achieve expansion planning through masking information included in the generator module of the framework. And experimental results based on various scales of power systems have shown the applicability of the proposed method. As for future studies, how to directly couple the resilience level with the design performance criteria remains to be a challenge. And the experimental results shown in this study only utilize a proxy for the system resilience, it's noteworthy to analyze more fundamental metrics to evaluate the designs during the generation.

%%%%% Acknowledgments %%%%%%%%%%%%%%%%%%%%%%%%%%%
\section*{Acknowledgments}
This research is partially supported the U.S. Department of Energy's Office of Nuclear Energy under Award No. DE-NE0008899 and the National Science Foundation (NSF) Engineering Research Center for Power Optimization of Electro-Thermal Systems (POETS) with cooperative agreement EEC-1449548.

%%%  REFERENCES  %%%%%%%%%%%%%%%%%%%%%%%%%%%%%%%%
%%
%% Put your references into your .bib file in the usual way. Run latex once, bibtex once, then latex twice.
%% The asmeconf.bst style allows: venue = {Location of Conference}, and eventdate = {Month, days}
%%		for @inproceedings and @proceedings
%%

\bibliographystyle{asmeconf}   %% .bst file following ASME conference format. Do not change.
\bibliography{asmeconf} %% <=== change this to name of your bib file

%%%  APPENDICES  %%%%%%%%%%%%%%%%%%%%%%%%%%%%%%%%
%\appendix

%%%%%%%%%%%%%%%%%%%%%%%%%%%%%%%%%%%%%%%%%%%%%%%%%%%%%%%%%%%%%%%%%%%%%%%%%%%%%%%%%%%%%%%

\end{document}